\pdfoutput=1

\documentclass[11pt]{article}

\usepackage[]{EMNLP2023}

\usepackage{times}
\usepackage{latexsym}
\usepackage{booktabs}
\usepackage{multirow}
\usepackage[T1]{fontenc}

\usepackage[utf8]{inputenc}

\usepackage{microtype}

\usepackage{wasysym}
\usepackage{inconsolata}
\usepackage{rotating}
\def\rot{\rotatebox}
\usepackage{algorithm}
\usepackage{algorithmic}

\usepackage{graphicx}
\usepackage{microtype}
\usepackage{enumitem}
\usepackage{inconsolata}
\usepackage{todonotes}
\usepackage[normalem]{ulem}
\usepackage{booktabs}
\usepackage{subcaption}
\usepackage{url}
\usepackage{graphicx}
\usepackage{multirow}
\usepackage{listings}
\usepackage{color}
\usepackage{pifont}
\newcommand{\cmark}{\ding{51}}%
\newcommand{\xmark}{\ding{55}}%

\newcommand{\blpsentiment}{BLP}
\newcommand{\rank}[1]{{\scriptsize \texttt{#1}}}

\title{BLP-2023 Task 2: Sentiment Analysis}

\author{
Md. Arid Hasan$^1$, Firoj Alam$^2$, Anika Anjum$^3$, Shudipta Das$^3$, Afiyat Anjum$^3$\\
$^1$SE+AI Research Lab, University of New Brunswick, Fredericton, Canada \\
$^2$Qatar Computing Research Institute, Doha, Qatar\\ 
$^3$Daffodil International University, Dhaka, Bangladesh\\
  \texttt{arid.hasan@unb.ca, fialam@hbku.edu.qa} \\
  }

\begin{document}
\maketitle

\begin{abstract}
We present an overview of the \blpsentiment{} Sentiment Shared Task, organized as part of the inaugural BLP 2023 workshop, co-located with EMNLP 2023. The task is defined as the detection of sentiment in a given piece of social media text. This task attracted interest from 71 participants, among whom 29 and 30 teams submitted systems during the development and evaluation phases, respectively. In total, participants submitted 597 runs. However, a total of 15 teams submitted system description papers. The range of approaches in the submitted systems spans from classical machine learning models, fine-tuning pre-trained models, to leveraging Large Language Model (LLMs) in zero- and few-shot settings. In this paper, we provide a detailed account of the task setup, including dataset development and evaluation setup. Additionally, we provide a brief overview of the systems submitted by the participants. All datasets and evaluation scripts from the shared task have been made publicly available for the research community, to foster further research in this domain.\footnote{\url{https://github.com/blp-workshop/blp_task2}}
\end{abstract}

\section{Introduction}
\label{sec:introduction}

Sentiment analysis has emerged as a significant sub-field in Natural Language Processing (NLP), with a wide array of applications encompassing social media monitoring, brand reputation management, market research, customer feedback analysis, among others. The advancement of sentiment analysis systems has been driven by substantial research efforts, addressing its indispensable utility across diverse fields such as business, finance, politics, education, and services \cite{cui2023survey}. Traditionally, analysis has been conducted across various types of content and domains including news articles, blog posts, customer reviews, and social media posts, and extended over different modalities like textual and multimodal analyses \cite{hussein2018survey,dashtipour2016multilingual}.

At its core, the task of sentiment analysis is defined as the extraction and identification of polarities (e.g., positive, neutral, and negative) expressed within texts. However, its scope has broadened to encompass the identification of: {\em(i)} the target (i.e., an entity) or aspect of the entity on which sentiment is expressed, {\em(ii)} the opinion holder, and {\em(iii)} the time at which it is expressed \cite{liu2020sentiment}. Such advancements have primarily been made for high-resource languages.

Research on fundamental sentiment analysis remains an ongoing exploration, especially for many low-resource languages, primarily due to the scarcity of datasets and consolidated community effort. Although there has been a recent surge in interest \cite{batanovic2016reliable,nabil2015astd,muhammad2023afrisenti}, the field continues to pose significant challenges. 
Similar to other low-resource languages, the challenges for sentiment analysis in Bangla have been reported in recent studies \cite{alam2021review,islam2021sentnob,islam2023sentigold}. \citet{alam2021review} emphasized the primary challenges associated with Bangla sentiment analysis, specifically issues of duplicate instances in the data, inadequate reporting of annotation agreement, and generalization. These challenges were also highlighted in \cite{islam2021sentnob}, further emphasizing the need to address them for effective sentiment analysis in Bangla.

To advance research in Bangla sentiment analysis, we emphasized community engagement and organized a shared task at BLP 2023. Similar efforts have primarily been conducted for other languages as part of the SemEval Workshop. The analysis of sentiment in tweets serves as an example of such efforts, particularly focusing on Arabic and English~\cite{rosenthal-etal-2017-semeval}. An earlier attempt at such an endeavor for Bangla is reported in \cite{patra2015shared}, which mainly focused on tweets. Our initiative significantly different from theirs in terms of datasets (e.g., data from multiple social media platforms and diverse domains) and evaluation setup.

A total of 71 teams registered for the task, out of which 30 made an official submission on the test set, and 15 of the participating teams submitted a system description paper.

The remainder of the paper is structured as follows: Section \ref{sec:related_work} provides an overview of the relevant literature. Section \ref{sec:dataset} discusses the task and dataset. Section \ref{sec:evaluation_framework} describes the organization of the task and the evaluation measures. An overview of the participating systems is provided in Section~\ref{sec:system_overview}. Lastly, Section \ref{sec:conclusion} concludes the paper.

\section{Related Work}
\label{sec:related_work}

The current state-of-the-art research for Bangla sentiment classification mainly dominated focuses on two key aspects: the development or datasets and model development. Notable recent work in this direction include \cite{6850712,alam2021review,islam2021sentnob, kabir2023banglabook,islam2023sentigold}. \citet{kabir2023banglabook} curated the largest dataset from book reviews, with annotations based on the review ratings. Although the dataset encompasses a large number of reviews, the class distribution poses a challenge for the Negative and Neutral classes. A well-balanced dataset has been explored in \cite{islam2021sentnob}, comprising $\sim$15K manually annotated comments spanning 13 different domains. This dataset is also used as a part of this shared task.

From a modeling perspective, the existing literature addresses the problem using both classical machine learning and deep learning algorithms. These include Naive Bayes, Support Vector Machine, Decision Tree, Maximum Entropy, and Random Forest \cite{9084046,banik2018evaluation,chowdhury2019analyzing, islam2016supervised}. Moreover, recent studies have extensively employed deep learning models for Bangla sentiment classification \cite{hassan2016sentiment,sharfuddin2018deep,tripto2018detecting,ashik2019data,DBLP:journals/corr/abs-2004-07807,sazzed2021improving,sharmin2021attention}. Common deep learning approaches incorporate LSTMs, CNNs, attention mechanisms, and multichannel convolutional LSTMs. In the studies by \citet{hasan2020sentiment, alam2021review}, comprehensive comparisons across various datasets were conducted, illustrating that the deep learning-based pretrained language model XLM-RoBERTa excels in performance. Comparisons between classical and deep learning-based approaches have also been explored \cite{ashik2019data,hasan2020sentiment,alam2021review}.

Given the significant capabilities that Large Language Models (LLMs) have demonstrated across diverse applications and scenarios, \citet{hasan2023zero} explored various LLMs such as Flan-T5 (large and XL) \cite{chung2022scaling}, Bloomz (1.7B, 3B, 7.1B, 176B-8bit) \cite{muennighoff2022crosslingual}, and GPT-4 \cite{openai2023gpt}, comparing the results with fine-tuned models. The resulting performance demonstrate that fine-tuned models continue to outperform zero- and few-shot prompting. However, the performance of LLMs showcases a promising direction towards the development of systems with limited datasets for new domains.

Though there is a surge of research interest and progress, utilizing such systems in real applications remains a challenge in terms of performance and generalization capability. This shared task aimed to advance research through community effort and focus on a standard evaluation setup. As a starting point, we aimed to classify sentiment into three sentiment polarities: positive, neutral, and negative. This approach can be further extended in future studies. 

\section{Task and Dataset}
\label{sec:dataset}

\subsection{Task}
The task is defined as ``detect the sentiment associated within a given text''. This is a multi-class classification task that involves determining whether the sentiment expressed in the text is \textit{Positive}, \textit{Negative}, and \textit{Neutral}.

\subsection{Dataset}
We utilized the MUBASE~\cite{hasan2023zero} and SentNoB~\cite{islam2021sentnob} datasets for the task. Both datasets were annotated by multiple annotators, with the inter-annotation agreement being $0.84$ for MUBASE and $0.53$ for SentNoB, respectively. The SentNoB data is curated from newspapers and YouTube video comments, covering 13 different topics such as Politics, National, International, Food, Sports, Teach, etc.
The MUBASE dataset consists of comments from popular news media sources such as BBC Bangla, Prothom Alo, and BD24Live, which were collected from Facebook and Twitter.

We further analyzed the distribution of sentences based on the number of words associated with each class label, as depicted in Table \ref{tab:detailed_data_dist}. We created various ranges of sentence length buckets to understand and define the sequence length while training the transformer-based models. It appears that more than 80\% of the posts comprise twenty words or fewer, a finding consistent with the typical of social media posts, as observed in previous studies \cite{alam2021humaid}. Moreover, the average number of words and sentences per data point are $15.87$ and $1.03$, respectively.

\begin{table}[!ht]
\centering
\begin{tabular}{llrrr}
\toprule
\multicolumn{1}{c}{\textbf{Split}} & \multicolumn{1}{c}{\textbf{\#Words}} & \multicolumn{1}{c}{\textbf{Pos}} & \multicolumn{1}{c}{\textbf{Neu}} & \multicolumn{1}{c}{\textbf{Neg}} \\ \midrule
\multirow{6}{*}{Train} & <10 & 5,616 & 3,595 & 6,575 \\
& 11-20 & 4,587 & 2,212 & 5,613\\
& 21-30 & 1,263 & 671 & 1,949\\
& 31-40 & 493 & 287 & 818\\ 
& 41-50 & 260 & 152 & 377\\
& 51+ & 145 & 218 & 435\\ \midrule
\multirow{6}{*}{Dev} & <10 & 587 & 398 & 723 \\
& 11-20 & 539 & 244 & 634\\
& 21-30 & 160 & 68 & 232\\
& 31-40 & 67 & 43 & 90\\ 
& 41-50 & 22 & 14 & 34\\
& 51+ & 13 & 26 & 40\\ \midrule
\multirow{6}{*}{Dev-test} & <10 & 601 & 292 & 783 \\
& 11-20 & 420 & 178 & 603\\
& 21-30 & 68 & 55 & 178\\
& 31-40 & 11 & 21 & 54\\ 
& 41-50 & 6 & 16 & 29\\
& 51+ & 20 & 38 & 53\\ \midrule
\multirow{6}{*}{Test} & <10 & 1,111 & 627 & 1,482 \\
& 11-20 & 762 & 382 & 1,183\\
& 21-30 & 140 & 121 & 371\\
& 31-40 & 31 & 56 & 111\\ 
& 41-50 & 16 & 26 & 71\\
& 51+ & 32 & 65 & 120\\ \bottomrule
\end{tabular}
\caption{Detailed class label distribution of the shared task data splits. Pos: Positive, Neu: Neutral, Neg: Negative.}
\label{tab:detailed_data_dist}
\end{table}

\begin{table}[!ht]
\centering
\setlength{\tabcolsep}{4pt}
\resizebox{0.38\textwidth}{!}{
\begin{tabular}{lrrrr}
\toprule
\multicolumn{1}{c}{\textbf{Dataset}} & \multicolumn{1}{c}{\textbf{Train}} & \multicolumn{1}{c}{\textbf{Dev}} & \multicolumn{1}{c}{\textbf{DT}} & \multicolumn{1}{c}{\textbf{Test}} \\ \midrule
MUBASE & \cmark & \xmark & \cmark & \cmark \\
SentNoB & \cmark & \cmark & \xmark & \xmark \\
\bottomrule
\end{tabular}
}
\caption{Data sources utilized in various splits for the shared task. DT: Dev-Test}
\label{tab:data_source}
\end{table}

For the shared task, we combined the MUBASE~\cite{hasan2023zero} training set with the SentNoB~\cite{islam2021sentnob} training set, resulting in a total of 35,266 entries for the training set. The SentNoB development set was used as the shared task development set. Additionally, the MUBASE development set served as the dev-test set for the shared task, while the test set was utilized for system evaluation and participant ranking. The specifics of the data sources are outlined in Table \ref{tab:data_source}, and the detailed distribution of the data split is presented in Table \ref{tab:data_dist}.

\begin{table}[!ht]
\centering
\setlength{\tabcolsep}{4pt}
\resizebox{0.5\textwidth}{!}{
\begin{tabular}{lrrrrr}
\toprule
\multicolumn{1}{c}{\textbf{Class}} & \multicolumn{1}{c}{\textbf{Train}} & \multicolumn{1}{c}{\textbf{Dev}} & \multicolumn{1}{c}{\textbf{DT}} & \multicolumn{1}{c}{\textbf{Test}} & \multicolumn{1}{c}{\textbf{Total}} \\ \midrule
Pos & 12,364 & 1,388 & 1,126 & 2,092 & 16,970 \\
Neu & 7,135 & 793 & 600 & 1,277 & 9,805 \\
Neg & 15,767 & 1,753 & 1,700 & 3,338 & 22,558 \\ \midrule
\textbf{Total} & \textbf{35,266} & \textbf{3,934} & \textbf{3,426} & \textbf{6,707} & \textbf{49,333} \\ \bottomrule
\end{tabular}
}
\caption{Class label distribution of the shared task dataset. DT: Dev-Test, Pos: Positive, Neu: Neutral, Neg: Negative}
\label{tab:data_dist}
\end{table}

\section{Evaluation Framework}
\label{sec:evaluation_framework}

\subsection{Evaluation Measures}
\label{ssec:evaluation_measures}
For evaluation, we used the \textit{Micro-F1 score} and the evaluation scripts along with data are available online\footnote{\url{https://github.com/blp-workshop/blp_task2}}. As reference points, we provided both the majority and random baselines. The majority baseline always predicts the most common class in the training data and assigns this class to each instance in the test dataset. Conversely, the random baseline assigns one of the classes randomly to each instance in the test dataset.

\subsection{Task Organization}
\label{task_organization}
For the shared task, we provided four sets of data: the training set, development set, development-test set, and test set, as outlined in Table \ref{tab:data_dist}. The purpose of providing the development set is for hyperparameter tuning. We provided the development test set without labels to allow participants to evaluate their systems during the system development phase. The test set was designated for the final system evaluation and ranking. We ran the shared task in two phases and hosted the submission system on the CodaLab platform.\footnote{\url{https://codalab.lisn.upsaclay.fr/competitions/14587}}

\paragraph{Development Phase} 
In the first phase, only the training set, development set, and development-test set were made available, with no gold labels provided for the latter. Participants competed against each other to achieve the best performance on the development test set. A live leaderboard was made available to keep track of all submissions. 

\paragraph{Test Phase} 
In the second phase, the test set was released without labels, and the participants were given just four days to submit their final predictions. The test set was used for evaluation and ranking. The leaderboard was set to private during the evaluation phase, and participants were allowed to submit multiple systems without seeing the scores. The last valid submission was considered for official ranking.

After the competition concluded, we released the test set with gold labels to enable participants to conduct further experiments and error analysis.
\section{Results and Overview of the Systems}
\label{sec:system_overview}

\subsection{Results}
\label{ssec:results}

A total of 29 and 30 teams submitted their systems during the development and evaluation phases, respectively. In Table \ref{tab:offical_rank}, we report the results of the submitted system on dev-test and test sets. We also include the results for the majority and random baselines. The ranking on the table was determined by the results from the test set. Note that some teams participated in the development phase but did not participate in the evaluation phase, and vice versa, as indicated by the symbol \xmark. Additionally, the team marked with $*$ did not submit a system description paper. 

Upon comparing the results from the dev-test and test sets across different teams, it appears that the performance difference between them is very minimal. The models did not exhibit overfitting; in some cases, the performance on the test set even surpassed that on the dev-test set.

As can be seen in Table \ref{tab:offical_rank}, almost all systems outperformed random baseline except one system, whereas 26 systems outperformed the majority baseline. The best system, Aambela~\cite{BLP2023:task2:Aambela}, achieved micro-F1 score of 0.73, which is an absolute improvement of 0.23. The team mainly fine-tuned BanglaBERT and multilingual BERT along with adversarial weight perturbation. The second best system, Knowdee~\cite{BLP2023:task2:knowdee}, used data augmentation with psudolabeling, which are obtained from an ensemble of models. The third best system, LowResource~\cite{BLP2023:task2:LowResource}, used ensemble of different fine-tuned models. 

In Table \ref{tab:overview_approaches}, we report the overview of the approaches of the submitted systems. 
The most used models are multilingual BERT, BanglaBERT, and XLM-RoBERTa. Specifically, 9, 8, and 14 out of 15 teams utilized multilingual BERT, BanglaBERT, and XLM-RoBERTa, respectively.
Ensembles of fine-tuned models provide the best systems for this task. Additionally, two teams applied few-shot learning using the mT5, BanglaBERT large, and GPT-3.5 models. However, the teams did not provide the details regarding the prompts.

\subsection{Discussion}
\label{ssec:discussion}
From the official ranking presented in Table \ref{tab:offical_rank}, 
early every team outperformed the performance of the random baseline system. The performance difference between the top 22 teams is very small compared with the 23rd-ranked team. In Table \ref{tab:top5_class_dist}, we presented the per-class performances for the top 5 teams. Although most of the teams performed better than the random baseline by a large margin, the neutral class is still the most difficult one to identify. The low performance in neutral class might be due to its skewed distribution in the dataset. Data augmentation, up-sampling the minority class, and class re-weighting are common approaches typically used to address such issues. Although some systems employed data augmentation, it seems this issue was not thoroughly considered across all teams.

\begin{table*}[!ht]
\centering
\setlength{\tabcolsep}{4pt}
\resizebox{0.65\textwidth}{!}{
\begin{tabular}{lrr}
\toprule
\multirow{2}{*}{\textbf{Rank -- Team}} & \multicolumn{2}{c}{\textbf{Micro-F1}} \\
 & \multicolumn{1}{c}{\textbf{Dev-Test}} & \multicolumn{1}{c}{\textbf{Test}}
\\ \midrule

1. Aambela~\cite{BLP2023:task2:Aambela} & 0.7303 & 0.7310 \\
2. Knowdee~\cite{BLP2023:task2:knowdee} & 0.7288 & 0.7267 \\
3. LowResource~\cite{BLP2023:task2:LowResource} & 0.7224 & 0.7179 \\
4. LowResourceNLU~\cite{BLP2023:task2:LowResourceNLU} & 0.7248 & 0.7172 \\
5. Z-Index~\cite{BLP2023:task2:z-index} & \xmark & 0.7164 \\
- ShadmanRohan* & 0.7207 & 0.7155 \\
6. RGB* & 0.7125 & 0.7112 \\
7. EmptyMind\cite{BLP2023:task2:emptymind} & 0.7215 & 0.7109 \\
8. KeAb* & 0.7125 & 0.7094 \\
9. Embeddings~\cite{BLP2023:task2:Embeddings} & \xmark & 0.7088 \\
10. RSM-NLP~\cite{BLP2023:task2:RSM-NLP} & 0.7023 & 0.7078 \\
11. DeepBlueAI* & \xmark & 0.7076 \\
12. nlpBDpatriots~\cite{BLP2023:task2:NLPBDPatriots} & 0.7192 & 0.7058 \\
13. NLP\_CUET* & 0.6278 & 0.7052 \\
14. M1437~\cite{BLP2023:task2:M1437} & 0.7315 & 0.7036 \\
15. Semantic\_Savants* & 0.6961 & 0.7002 \\
16. meemaw* & \xmark & 0.6996 \\
17. Score\_IsAll\_You\_Need* & 0.6909 & 0.6930 \\
18. VishwasGPai* & 0.6970 & 0.6824 \\
19. UFAL-ULD~\cite{BLP2023:task2:souro} & 0.6661 & 0.6768 \\
20. Semantics Squad~\cite{BLP2023:task2:SemanticsSquad} & 0.7201 & 0.6742 \\
21. BanglaNLP~\cite{BLP2023:task2:BanglaNLP} & 0.6810 & 0.6702 \\
22. VacLM* & \xmark & 0.6584 \\
23. trina* &\xmark & 0.6194 \\
- Rachana8.\_K* & \xmark & 0.5962 \\
24. lixn* & \xmark & 0.5889 \\
25. Baseline (Majority) & 0.4962 & 0.4977 \\
26. Xenon* & \xmark & 0.4534 \\
27. Error Point~\cite{BLP2023:task2:ErrorPoing} & \xmark & 0.4129 \\
28. SSCP* & 0.5584 & 0.3390 \\
29. Baseline (Random) & 0.3389 & 0.3356 \\
30. Ushoshi2023~\cite{BLP2023:task2:ushoshi} & \xmark & 0.2626 \\ 
-- Shilpa* & 0.7166 & \xmark\\
-- Dhiman* & 0.7154 & \xmark\\
-- KarbonDark* & 0.7154 & \xmark\\
-- MrinmoyMahato* & 0.7107 & \xmark\\
-- shakib034* & 0.6734 & \xmark\\
-- Saumajit* & 0.6559 & \xmark\\
-- sankalok* & 0.6203 & \xmark\\
-- DiscoDancer420* & 0.5736 & \xmark\\
-- Devs* & 0.5736 & \xmark\\
-- almamunsardar* & 0.5642 & \xmark\\
\bottomrule
\end{tabular}
}
\caption{Official ranking of the shared task on the test set. *No working note submitted. - Run submitted after the deadline. 
\xmark~- indicates team has not submitted system in the respective phase.
}
\label{tab:offical_rank}
\end{table*}

\begin{table*}[!ht]
    \centering
\resizebox{\textwidth}{!}{
    \begin{tabular}{l|ccccccccccccccccc|cc}
    \toprule
    \multicolumn{1}{c}{\textbf{Team}} & \multicolumn{17}{c}{\textbf{Models}} & \multicolumn{2}{c}{\textbf{Misc.}} \\
    & \rot{90}{\textbf{Classical}} & \rot{90}{\textbf{multilingual BERT}} & \rot{90}{\textbf{RoBERTa}}&\rot{90}{\textbf{XLM-RoBERTa}}&\rot{90}{\textbf{BanglaBERT}}&\rot{90}{\textbf{BanglishBERT}}&\rot{90}{\textbf{MuRIL}}&\rot{90}{\textbf{mT5}}&\rot{90}{\textbf{BanglaT5}}&\rot{90}{\textbf{Indic-BERT}}&\rot{90}{\textbf{BanglaGPT2}}&\rot{90}{\textbf{DistilBERT}}&\rot{90}{\textbf{LSTM}}&\rot{90}{\textbf{LSTM-CNN}}&\rot{90}{\textbf{Bangla-BERT}}&\rot{90}{\textbf{Few-shot}}&\rot{90}{\textbf{Ensemble}}&\rot{90}{\textbf{Preprocessing}}&\rot{90}{\textbf{Data Augmentation}} \\ \midrule
    Aambela~\cite{BLP2023:task2:Aambela} & & \cmark & & & \cmark & & & & & & & & & & & & \cmark & \cmark& \\
    Knowdee~\cite{BLP2023:task2:knowdee} & & & & \cmark & \cmark & & \cmark & \cmark & & & & & & & & \cmark & \cmark & \cmark & \cmark \\
    LowResource~\cite{BLP2023:task2:LowResource} & & & & \cmark & \cmark & & & & \cmark & & & & & & & & \cmark & & \\
    LowResourceNLU~\cite{BLP2023:task2:LowResourceNLU} & & \cmark & & & \cmark & & & & & & & & & & & & \cmark & \cmark & \\
    Z-Index~\cite{BLP2023:task2:z-index} & \cmark & \cmark & & & \cmark & & & & & & & & & & & & & \cmark & \\ 
    EmptyMind~\cite{BLP2023:task2:emptymind} & \cmark & & & & \cmark & & & & & & & & \cmark & & & & & & \\
    Embeddings~\cite{BLP2023:task2:Embeddings} & & \cmark & & & \cmark & & & & & \cmark & \cmark & & & & & & & & \\ 
    RSM-NLP~\cite{BLP2023:task2:RSM-NLP} & & & \cmark & & \cmark & \cmark & \cmark & & & & & \cmark & & & \cmark & & \cmark & \cmark & \cmark \\ 
    nlpBDpatriots~\cite{BLP2023:task2:NLPBDPatriots} & \cmark & \cmark & & \cmark & \cmark & & \cmark & & & & & & & & & \cmark & & & \\ 
    M1437~\cite{BLP2023:task2:M1437} & & & & \cmark & \cmark & & & & & & & & & & & & & & \\ 
    UFAL-ULD~\cite{BLP2023:task2:souro} & & \cmark & & \cmark & \cmark & & & & & & & & & & \cmark & & & \cmark & \cmark \\ 
    Semantics Squad~\cite{BLP2023:task2:SemanticsSquad} & & \cmark & & \cmark & \cmark & \cmark & & & & & & & & & & & & \cmark & \\
    BanglaNLP~\cite{BLP2023:task2:BanglaNLP} & \cmark & \cmark & & \cmark & \cmark & & & & & & & & & & & & & & \\
    Error Point~\cite{BLP2023:task2:ErrorPoing} & \cmark & & & & & & & & & & & & \cmark & \cmark & & & & \cmark & \cmark \\
    Ushoshi2023~\cite{BLP2023:task2:ushoshi} & \cmark & \cmark & & \cmark & \cmark & & & & & & & \cmark & \cmark & & & & & \cmark & \cmark \\ \bottomrule
    \end{tabular}
    }
    \caption{Overview of the approaches used in the submitted systems.}
    \label{tab:overview_approaches}
\end{table*}

\subsection{Participating Systems}
\label{ssec:system_description}
Below, we provide a brief description of the participating systems and their leaderboard rank. 

\textbf{Aambela~\cite{BLP2023:task2:Aambela} (\rank{rank 1})} emerged as the best-performing team in the shared task, fine-tuning pretrained models BanglaBERT~\cite{bhattacharjee-etal-2022-banglabert} and multilingual BERT~\cite{devlin2018bert} using two classification heads. Initially, the author removed URLs and HTML tags, then applied a normalizer to the preprocessed text. Adversarial weight perturbation was utilized to enhance the training's robustness, and a 5-fold cross-validation was also conducted.

\textbf{Knowdee~\cite{BLP2023:task2:knowdee} (\rank{rank 2})} partitioned the data set into 10 folds and generated pseudo-labels for unlabeled data using a fine-tuned ensemble of models. They employed standard data preprocessing and augmentation techniques to process the data, and fine-tuned BanglaBERT~\cite{bhattacharjee-etal-2022-banglabert}, MuRIL~\cite{khanuja2021muril}, XLM-RoBERTa~\cite{conneau2019unsupervised}, and mT5~\cite{xue2020mt5}, achieving the second-best performance. The team also implemented Few-shot (3-shot) learning and compared the results with those from fine-tuned models.

\textbf{LowResource~\cite{BLP2023:task2:LowResource} (\rank{rank 3})} fine-tuned both the base and large versions of BanglaBERT~\cite{bhattacharjee-etal-2022-banglabert}, employing randomly dropping tokens, and also fine-tuned XLM-RoBERTa~\cite{conneau2019unsupervised}. During the development phase, they created an ensemble of three models. However, for the evaluation phase, they ensembled only two variants of BanglaBERT, with one of them being fine-tuned using external data. Additionally, they employed task-adaptive pretraining and paraphrasing techniques utilizing BanglaT5~\cite{bhattacharjee2022banglanlg}.

\begin{table*}[!ht]
    \centering
    \begin{tabular}{lr|rrrrr}
    \toprule
    \textbf{Class} & \textbf{Baseline} & \textbf{Aambela} & \textbf{Knowdee} & \textbf{LowResource} & \textbf{LowResourceNLU} & \textbf{Z-Index} \\ \midrule
    Negative & $0.3996$ & $0.7958$ & $0.7943$ & $0.7873$ & $0.7877$ & $0.7877$\\
    Neutral & $0.2368$ & $0.4998$ & $0.4592$ & $0.3998$ & $0.4021$ & $0.4250$\\
    Positive & $0.3329$ & $0.7666$ & $0.7599$ & $0.7567$ & $0.7530$ & $0.7559$\\ 
    \bottomrule
    \end{tabular}
    \caption{F1 scores of the baseline and top five systems for each class.}
    \label{tab:top5_class_dist}
\end{table*}

\textbf{LowResourceNLU~\cite{BLP2023:task2:LowResourceNLU} (\rank{rank 4})} fine-tuned BanglaBERT base and large ~\cite{bhattacharjee-etal-2022-banglabert}, with MLM and classification heads, and multilingual BERT~\cite{devlin2018bert} jointly on the XNLI and shared task dataset. They also created an ensemble of all three transformer-based models and applied multi-step aggregation to capture the most confident class predicted across all models.

\textbf{Z-Index~\cite{BLP2023:task2:z-index} (\rank{rank 5})} utilized standard preprocessing techniques to remove URLs, usernames, emojis, and hashtags from the text. Initially, they employed SVM and Random Forest classical models, and later fine-tuned both the base and large variants of BanglaBERT~\cite{bhattacharjee-etal-2022-banglabert}, as well as the multilingual BERT~\cite{devlin2018bert}. The model was trained using the provided training set.

\textbf{EmptyMind~\cite{BLP2023:task2:emptymind} (\rank{rank 7})} initially applied classical models such as Decision Tree, Random Forest, SVM, and XGBoost, utilizing TF-IDF vectors, as well as Word2Vec vectors. Subsequently, they employed deep learning-based models including Stacked BiLSTM and BiLSTM+CNN. Furthermore, they fine-tuned different variants of BanglaBERT~\cite{bhattacharjee-etal-2022-banglabert}.

\textbf{Embeddings~\cite{BLP2023:task2:Embeddings} (\rank{rank 9})} fine-tuned pretrained models BanglaBERT~\cite{bhattacharjee-etal-2022-banglabert}, BanglaGPT2,\footnote{\url{https://huggingface.co/flax-community/gpt2-bengali}} 
Indic-BERT~\cite{kakwani2020indicnlpsuite}, and multilingual BERT\cite{devlin2018bert} using cross entropy loss function. Later to reduce the computational cost, they investigated the performances across the self-adjusting dice loss, focal loss, and F1-micro loss. They also combined training, dev, and dev-test sets as training data to train and test data to evaluate the performances of the models. 

\textbf{RSM-NLP~\cite{BLP2023:task2:RSM-NLP} (\rank{rank 10})} submitted their runs by fine-tuning RoBERTa~\cite{liu2019roberta}, DistilBERT~\cite{sanh2019distilbert}, Bangla-BERT,\footnote{\url{https://github.com/sagorbrur/bangla-bert}} 
BanglaBERT~\cite{bhattacharjee-etal-2022-banglabert}, BanglishBERT~\cite{bhattacharjee-etal-2022-banglabert}, and MuRIL~\cite{khanuja2021muril}, with the additional use of training data. They employed standard preprocessing techniques to process the data. They also submitted ensemble techniques (i.e., weighted and majority-voted) of fine-tuned models.

\textbf{nlpBDpatriots~\cite{BLP2023:task2:NLPBDPatriots} (\rank{rank 12})} began with traditional approaches such as logistic regression and SVM. Later, they fine-tuned BanglaBERT~\cite{bhattacharjee-etal-2022-banglabert}, multilingual BERT~\cite{devlin2018bert}, MuRIL~\cite{khanuja2021muril}, and XLM-RoBERTa~\cite{conneau2019unsupervised}, and ensemble the models using a weighted average of the confidence predicted by each model. They also employed few-shot learning using GPT-3.5~\cite{openai2023gpt}.

\textbf{M1437~\cite{BLP2023:task2:M1437} (\rank{rank 14})} fine-tuned large pretrained language models BanglaBERT large~\cite{bhattacharjee-etal-2022-banglabert} and XLM-RoBERTa large~\cite{conneau2019unsupervised} along with the base version of each model. They also used an existing dataset~\cite{hasan2020sentiment} in addition to the provided training data. To compare among the transformers models, they also fine-tuned the multilingual BERT. During the development phase, they were the best-performing team and they ended the competition in the $14{th}$ position in the evaluation phase.

\textbf{UFAL-ULD~\cite{BLP2023:task2:souro} (\rank{rank 19})} fine-tuned BanglaBERT~\cite{bhattacharjee-etal-2022-banglabert}, Bangla-BERT\footnote{\url{https://github.com/sagorbrur/bangla-bert}} 
multilingual BERT~\cite{devlin2018bert}, and XLM-RoBERTa~\cite{conneau2019unsupervised} to tackle the problem. They followed the standard preprocessing steps to process the data and upsampled the training data to achieve balance among the classes. They also employed a focal loss function to address hard-to-classify examples.

\textbf{Semantics Squad~\cite{BLP2023:task2:SemanticsSquad} (\rank{rank 20})} submitted runs for both the development and evaluation phases. Standard preprocessing techniques were applied, with URLs and hashtags being removed from the data, to process and fine-tune BanglaBERT~\cite{bhattacharjee-etal-2022-banglabert}, BanglishBERT~\cite{bhattacharjee-etal-2022-banglabert}, XLM-RoBERTa~\cite{conneau2019unsupervised}, and multilingual BERT\cite{devlin2018bert}.

\textbf{BanglaNLP~\cite{BLP2023:task2:BanglaNLP} (\rank{rank 21})} also fine-tuned BanglaBERT~\cite{bhattacharjee-etal-2022-banglabert}, BERT multilingual~\cite{devlin2018bert}, and XLM-RoBERTa~\cite{conneau2019unsupervised} pretrained models. Additionally, they performed parameter-efficient tuning (P-tuning) on XLM-RoBERTa. They also employed traditional models such as Logistic Regression, Naive Bayes, SGD Classifier, Majority Voting, and Stacking in their approach to the task.

\textbf{Error Point~\cite{BLP2023:task2:ErrorPoing} (\rank{rank 27})} performed preprocessing by removing duplicate text, filtering based on text length, and eliminating punctuation, links, emojis, non-character elements, and stopwords. They also carried out data augmentation. For their analysis, they utilized classical algorithms such as Logistic Regression, Decision Tree, Random Forest, Multinomial Naive Bayes, SVM, and SGD, using $n$-grams to represent the input. Additionally, they employed deep learning models, namely LSTM and LSTM-CNN.

\textbf{Ushoshi2023~\cite{BLP2023:task2:ushoshi} (\rank{rank 30})} applied preprocessing by removing punctuation marks, links, emojis, hashtag signs, usernames, and non-Bangla characters. They also applied an upsampling technique to balance the dataset. Initially, they employed traditional models such as logistic regression, decision tree, random forest, multinomial naive bayes, k-nearest neighbor, SVM, and SGD for classification. Subsequently, they fine-tuned BanglaBERT~\cite{bhattacharjee-etal-2022-banglabert}, XLM-RoBERTa~\cite{conneau2019unsupervised}, DistilBERT~\cite{sanh2019distilbert}, and multilingual BERT~\cite{devlin2018bert}. Additionally, they trained a deep learning model, LSTM, to compare the performances across different models.

\section{Conclusion and Future Work}
\label{sec:conclusion}

We presented an overview of the shared task 2 (sentiment analysis) at the BLP Workshop 2023. Task 2 aimed to classify the sentiment in textual content. Notable systems employed an ensemble of pretrained language models, with the language-specific BanglaBERT being the most popular. Also, some interesting approaches including P-tuning, Few-shot learning, LLMs, and different loss functions have been explored for tackling the problem. In general, numerous models, including different kinds of transformers, have been used in the current submissions for the task. 

In future work, we plan to extend the task in various ways, such as aspect-based sentiment analysis and incorporating multiple modalities.

\section*{Limitations}
The BLP-2023 sentiment analysis shared task primarily focuses on sentiment polarity classification (positive, negative, and neutral) at the post level. This approach limits the identification of specific sentiment aspects and other crucial elements associated with them. Future editions of the task will address this aspect. Moreover, this edition focused solely on unimodality (text-only) models, leaving multimodal models for future study.

\bibliographystyle{acl_natbib}
\bibliography{bib/anthology, bib/aaai}




\end{document}